\documentclass[runningheads]{llncs}
\usepackage{graphicx}
   
\usepackage{amsfonts}
\usepackage{bbm}
\usepackage[colorlinks=true,linkcolor=blue, citecolor=blue, urlcolor=blue]{hyperref}
\usepackage{hhline}
\usepackage{ulem}
\usepackage[table]{xcolor}
\usepackage{stmaryrd}
\usepackage{amsmath}
\usepackage[misc]{ifsym}

\DeclareMathSymbol{\shortminus}{\mathbin}{AMSa}{"39}

\usepackage{fdsymbol}

\begin{document}
\title{COLosSAL: A Benchmark for Cold-start Active Learning for 3D Medical Image Segmentation}
\titlerunning{Cold-start AL for 3D medical image segmentation}
\author{Han Liu\inst{1}\Letter\and
Hao Li\inst{1} \and
Xing Yao\inst{1} \and
Yubo Fan\inst{1} \and
Dewei Hu\inst{1} \and
Benoit Dawant\inst{1} \and
Vishwesh Nath\inst{2} \and
Zhoubing Xu\inst{3} \and
Ipek Oguz\inst{1}}

\institute{
Vanderbilt University\and
NVIDIA\and 
Siemens Healthineers
\\
\email{han.liu@vanderbilt.edu}}
\authorrunning{H. Liu et al.}

\maketitle              
\begin{abstract}
Medical image segmentation is a critical task in medical image analysis. In recent years, deep learning based approaches have shown exceptional performance when trained on a fully-annotated dataset. However, data annotation is often a significant bottleneck, especially for 3D medical images. Active learning (AL) is a promising solution for efficient annotation but requires an initial set of labeled samples to start active selection. When the entire data pool is unlabeled, how do we select the samples to annotate as our initial set? This is also known as the cold-start AL, which permits only one chance to request annotations from experts without access to previously annotated data. Cold-start AL is highly relevant in many practical scenarios but has been under-explored, especially for 3D medical segmentation tasks requiring substantial annotation effort. In this paper, we present a benchmark named COLosSAL by evaluating six cold-start AL strategies on five 3D medical image segmentation tasks from the public Medical Segmentation Decathlon collection. We perform a thorough performance analysis and explore important open questions for cold-start AL, such as the impact of budget on different strategies. Our results show that cold-start AL is still an unsolved problem for 3D segmentation tasks but some important trends have been observed. The code repository, data partitions, and baseline results for the complete benchmark are publicly available at \url{https://github.com/MedICL-VU/COLosSAL}.

\end{abstract}

\section{Introduction}
Segmentation is among the most common medical image analysis tasks and is critical to a wide variety of clinical applications. To date, data-driven deep learning (DL) methods have shown prominent segmentation performance when trained on fully-annotated datasets \cite{hesamian2019deep}. However, data annotation is a significant bottleneck for dataset creation. First, annotation process is tedious, laborious and time-consuming, especially for 3D medical images where dense annotation with voxel-level accuracy is required. Second, medical images typically need to be annotated by medical experts whose time is limited and expensive, making the annotations even more difficult and costly to obtain. Active learning (AL) is a promising solution to improve annotation efficiency by iteratively selecting the most \textit{important} data to annotate with the goal of reducing the total number of annotated samples required. However, most deep AL methods require an initial set of labeled samples to start the active selection. When the entire data pool is unlabeled, which samples should one select as the initial set? This problem is known as \textbf{\textit{cold-start active learning}}, a low-budget paradigm of AL that permits only one chance to request annotations from experts without access to any previously annotated data. 

Cold-start AL is highly relevant to many practical scenarios. First, cold-start AL aims to study the general question of constructing a training set for an organ that has not been labeled in public datasets. This is a very common scenario (whenever a dataset is collected for a new application), especially when iterative AL is not an option. Second, even if iterative AL is possible, a better initial set  has been found to lead to noticeable improvement for the subsequent AL cycles \cite{chen2023making,zheng2019biomedical}. Third, in low-budget scenarios, cold-start AL can achieve one-shot selection of the most informative data without several cycles of annotation. This can lead to an appealing \textit{`less is more'} outcome by optimizing the available budget and also alleviating the issue of having human experts on standby for traditional iterative AL.

Despite its importance, very little effort has been made to address the cold-start problem, especially in medical imaging settings. The existing cold-start AL techniques are mainly based on the two principles of the traditional AL strategies: (1) \textit{Uncertainty sampling} \cite{ranganathan2017deep,lewis1994heterogeneous,nguyen2022measure,gaillochet2022taal}, where the most uncertain samples are selected to maximize the added value of the new annotations. (2) \textit{Diversity sampling} \cite{yuan2020cold,jin2022cold,sener2017active,hacohen2022active}, where samples from diverse regions of the data distribution are selected to avoid redundancy. In the medical domain, diversity-based cold-start strategies have been recently explored on 2D classification/segmentation tasks \cite{chen2023making,zhao2022self,zheng2019biomedical}. The effectiveness of these approaches on 3D medical image segmentation remains unknown, especially since 3D models are often patch-based while 2D models can use the entire image. A recent study on 3D medical segmentation shows the feasibility to use the uncertainty estimated from a proxy task to rank the importance of the unlabeled data in the cold-start scenario \cite{nath2022warm}. However, it fails to compare against the diversity-based approaches, and the proposed proxy task is only limited to CT images, making the effectiveness of this strategy unclear on other 3D imaging modalities. Consequently, no comprehensive cold-start AL baselines currently exist for 3D medical image segmentation, creating additional challenges for this promising research direction.

In this paper, we introduce the COLosSAL benchmark, the first \underline{col}d-\underline{s}tart \underline{a}ctive \underline{l}earning benchmark for 3D medical image segmentation by evaluating on six popular cold-start AL strategies. Specifically, we aim to answer three important open questions: (1) compared to random selection, how effective are the uncertainty-based and diversity-based cold-start strategies for 3D segmentation tasks? (2) what is the impact of allowing a larger budget on the compared strategies? (3) can these strategies work better if the local ROI of the target organ is known as prior? We train and validate our models on five 3D medical image segmentation tasks from the publicly available Medical Segmentation Decathlon (MSD) dataset \cite{antonelli2022medical}, which covers two of the most common 3D image modalities and the segmentation tasks for both healthy tissue and tumor/pathology.

Our contributions are summarized as follows:
\begin{itemize}
    \item We offer the first cold-start AL benchmark for 3D medical image segmentation. We make our code repository, data partitions, and baseline results publicly available to facilitate future cold-start AL research.
    \item We explore the impact of the budget and the extent of the 3D ROI on the cold-start AL strategies.
    \item Our major findings are: (1) TypiClust \cite{hacohen2022active}, a diversity-based approach, is a more robust cold-start selection strategy for 3D segmentation tasks. (2) Most evaluated strategies become more effective when more budget is allowed, especially diversity-based ones. (3) Cold-start AL strategies that focus on the uncertainty/diversity from a local ROI cannot outperform their global counterparts. (4) Almost no cold-start AL strategy is very effective for the segmentation tasks that include tumors.
\end{itemize}

\section{COLosSAL Benchmark Definition}
Formally, given an unlabeled data pool of size $N$, cold-start AL aims to select the optimal $m$ samples $(m\ll{N})$  \textit{without} access to any prior segmentation labels. Specifically, the optimal samples are defined as the subset of 3D volumes that can lead to the best validation performance when training a standard 3D segmentation network. In this study, we use $m=5$ for low-budget scenarios.

\subsection{3D Medical Image Datasets}
We use the Medical Segmentation Decathlon (MSD) collection \cite{antonelli2022medical} to define our benchmark, due to its public accessibility and the standardized datasets spanning across two common 3D image modalities, i.e., CT and MRI. We select five tasks from the collection appropriate for the 3D segmentation tasks, namely tasks 2-Heart, 3-Liver, 4-Hippocampus, 7-Pancreas, and 9-Spleen. Liver and Pancreas tasks include both organ and tumor segmentation, while the other tasks focus on organs  only. The selected tasks thus include  different organs with different disease status, representing a good coverage of real-world 3D medical image segmentation tasks. For each dataset, we split the data into training and validation sets for AL development. The training and validation sets contain 16/4 (heart), 105/26 (hippocampus), 208/52 (liver), 225/56 (pancreas), and 25/7 (spleen) subjects. The training set is considered as the unlabeled data pool for sample selection, and the validation set is kept consistent for all experiments to evaluate the performance of the selected samples by different AL schemes. 

\begin{figure}[t]
\includegraphics[width=\textwidth]{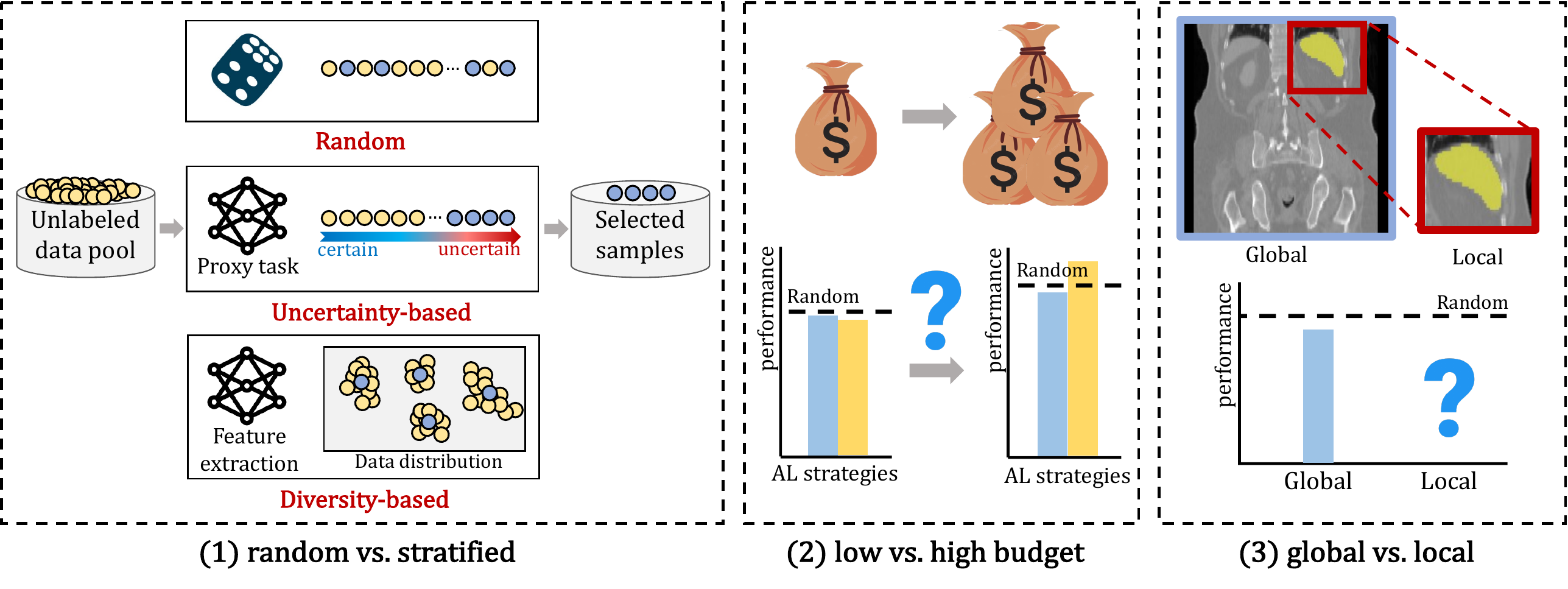}
\centering
\caption{Illustration of our three cold-start AL scenarios. We evaluate \textbf{(1)} uncertainty and diversity based selection strategies against random selection in a low-budget regime, \textbf{(2)} the effect of budget on performance, and \textbf{(3)} the usefulness of a local ROI for selection strategies.}
\label{fig1}
\end{figure} 

\subsection{Cold-start AL Scenarios}
In this study, we investigate the cold-start AL strategies for 3D segmentation tasks in three scenarios, as illustrated in Fig. \ref{fig1}. 

\begin{enumerate}
    \item With a low budget of 5 volumes (except for Heart, where 3 volumes are used because of the smaller dataset and easier segmentation task), we assess the performance of the uncertainty-based and diversity-based approaches against the random selection. 
    \item Next, we explore the impact of budgets for different cold-start AL schemes by allowing a higher budget, as previous work shows inconsistent effectiveness of AL schemes in different budget regimes \cite{hacohen2022active}. 
    \item Finally, we explore whether the cold-start AL strategies can benefit from using the uncertainty/diversity from  only the local ROI of the target organ, rather than the entire volume. This strategy may be helpful for 3D tasks especially for small organs, whose uncertainty/diversity can be outweighted by the irrelevant structures in the entire volume, but needs to be validated. 
\end{enumerate}
\textbf{Evaluation Metrics.} To evaluate the segmentation performance, we use the Dice similarity coefficient and 95\% Hausdorff distance (HD95), which measures the overlap between the segmentation result and ground truth, and the quality of segmentation boundaries by computing the $95^\mathrm{th}$ percentile of the distances between the segmentation and the ground truth boundary points, respectively.

\subsection{Baseline Cold-start Active Learners}
We provide the implementation for the baseline approaches: random selection, two variants of an uncertainty-based approach named ProxyRank \cite{nath2022warm}, and three diversity-based methods, namely ALPS \cite{yuan2020cold}, CALR \cite{jin2022cold}, and TypiClust \cite{hacohen2022active}. 

\noindent
\textbf{\underline{Random Selection.}} As suggested by prior works \cite{chen2023making,hacohen2022active,mittal2019parting,zhu2019addressing,simeoni2021rethinking,chandra2021initial}, random selection is a strong competitor in the cold-start setting, since it is independent and identically distributed (i.i.d.) to the entire data pool. We shuffle the entire training list with a random seed and select the first $m$ samples. In our experiments, random selection is conducted 15 times and the mean Dice score is reported.

\noindent
\textbf{\underline{Uncertainty-based Selection.}} Many traditional AL methods use uncertainty sampling, where the most uncertain samples are selected using the uncertainty of the network trained on an initial labeled set. Without such an initial labeled set, it is not straightforward to capture uncertainty in the cold-start setting. 

Recently, Nath \textit{et al.} \cite{nath2022warm} proposed a proxy task and then utilized uncertainty generated from the proxy task to rank the unlabeled data. By selecting the most uncertain samples, this strategy has shown superior performance  to random selection. 
Specifically, pseudo labels were generated by thresholding the CT images with an organ-dependent Hounsfield Unit (HU) intensity window. These pseudo labels carry coarse information for the target organ, though they also include other unrelated structures. The uncertainty generated by this proxy task is assumed to represent the uncertainty of the actual segmentation task. 

However, this approach \cite{nath2022warm} was limited to CT images. Here, we extend this strategy to MR images. For each MR image, we apply a sequence of transformations to convert it to a noisy binary mask: (1) z-score normalization, (2) intensity clipping to the [$1^\mathrm{st}, 99^\mathrm{th}$] percentile of the intensity values, (3) intensity normalization to [$0, 1$] and (4) Otsu thresholding \cite{otsu1979threshold}. We visually verify that the binary pseudo label includes the coarse boundary of the target organ.

As in \cite{nath2022warm}, we compute the model uncertainty for each unlabeled data using Monte Carlo dropout \cite{gal2016dropout}: with dropout enabled during inference, multiple predictions are generated with stochastic dropout configurations. Entropy \cite{nath2020diminishing} and Variance \cite{yang2017suggestive} are used as uncertainty measures to create two variants of this proxy ranking method, denoted as \textbf{ProxyRank-Ent} and \textbf{ProxyRank-Var}. The overall uncertainty score of an unlabeled image is computed as the mean across all voxels. Finally, we rank all unlabeled data with the overall uncertainty scores and select the most uncertain $m$ samples.

\noindent
\textbf{\underline{Diversity-based Selection.}} Unlike uncertainty-based methods which require a warm start, diversity-based methods can be used in the cold-start setting. Generally, diversity-based approaches consist of two stages. First, a feature extraction network is trained using unsupervised/self-supervised tasks to represent each unlabeled data as a latent feature. Second, clustering algorithms are used to select the most diverse samples in latent space to reduce data redundancy. The major challenge of benchmarking the diversity-based methods for 3D tasks is to have a feature extraction network for 3D volumes. To address this issue, we train a 3D auto-encoder on the unlabeled training data using a self-supervised task, i.e., image reconstruction. Specifically, we represent each unlabeled 3D volume as a latent feature by extracting the bottleneck feature maps, followed by an adaptive average pooling for dimension reduction \cite{zhao2022self}.

Afterwards, we adapt the diversity-based approaches to our 3D tasks by using the same clustering strategies as proposed in the original works, but replacing the feature extraction network with our 3D version. In our benchmark, we evaluate the clustering strategies from three state-of-the-art diversity-based methods. \begin{enumerate}
    \item \textbf{ALPS} \cite{yuan2020cold}: \textit{k}-MEANS is used to cluster the latent features with the number of clusters equal to the query number $m$. For each cluster, the sample that is the closest to the cluster center is selected. 
    \item \textbf{CALR} \cite{jin2022cold}: This approach is based on the maximum density sampling, where the sample with the most information is considered the one that can optimally represent the distribution of a cluster. A bottom-up hierarchical clustering algorithm termed BIRCH \cite{zhang1996birch} is used and the number of clusters is set as the query number $m$. For each cluster, the information density for each sample within the cluster is computed and the sample with the highest information density is selected. The information density is expressed as $I(x)=\frac{1}{|X_{c}|}{\sum_{x'\in{X_{c}}}{sim(x,x')}}$, where $X_{c}=\{x_{1},x_{2},...x_{j}\}$ is the feature set in a cluster and cosine similarity is used as $sim(\cdot)$.
    \item \textbf{TypiClust} \cite{hacohen2022active}: This approach also uses the points density in each cluster to select a diverse set of typical examples. \textit{k}-MEANS clustering is used, followed by selecting the most typical data from each cluster, which is similar to the ALPS strategy but less sensitive to outliers. The typicality is calculated as the inverse of the average Euclidean distance of $x$ to its $K$ nearest neighbors KNN$(x)$, expressed as: $\mathrm{Typicality}(x)=(\frac{1}{K}{\sum_{x_{i}\in{\mathrm{KNN}(x)}}{||x-x_{i}||_{2}})^{-1}}$. $K$ is set as 20 in the original paper but that is too high for our application. Instead, we use all the samples from the same cluster to calculate typicality. 
\end{enumerate} 

\subsection{Implementation Details}
In our benchmark, we use the 3D U-Net as the network architecture. For uncertainty estimation, 20 Monte Carlo simulations are used with a dropout rate of $0.2$. As in \cite{nath2022warm}, a dropout layer is added at the end of every level of the U-Net for both encoder and decoder. The performance of different AL strategies is evaluated by training a 3D patch-based segmentation network using the selected data, which is an important distinction from the earlier 2D variants in the literature. The only difference between different experiments is the selected data. For CT pre-processing, image intensity is clipped to [$-1024, 1024$] HU and rescaled to [$0,1$]. For MRI pre-processing, we sequentially apply z-score normalization, intensity clipping to [$1^\mathrm{st}, 99^\mathrm{th}$] percentile and rescaling to [$0,1$]. 
During training, we randomly crop a 3D patch with a patch size of $128 \times{} 128 \times{} 128$ (except for hippocampus, where we use $32 \times{} 32 \times{} 32$) with the center voxel of the patch being foreground and background at a ratio of $2:1$. Stochastic gradient descent algorithm with a Nesterov momentum ($\mu=0.99$) is used as the optimizer and $L_\mathrm{DiceCE}$ is used as the segmentation loss. An initial learning rate is set as $0.01$ and decayed with a polynomial policy as in \cite{isensee2021nnu}. For each experiment, we train our model using 30k iterations and validate the performance every 200 iterations. A variety of augmentation techniques as in \cite{isensee2021nnu} are applied to achieve optimal performance for all compared methods. All the networks are implemented in PyTorch \cite{paszke2019pytorch} and MONAI \cite{cardoso2022monai}. Our experiments are conducted with the deterministic training mode in MONAI with a fixed random seed=0. We use a 24G NVIDIA GeForce RTX 3090 GPU. 

For the global vs.\ local experiments, the local ROIs are created by extracting the 3D bounding box from the ground truth mask and  expanding it by five voxels along each direction. We note that although no ground truth masks are accessible in the cold-start AL setting, this analysis is still valuable to determine the usefulness of local ROIs. It is only worth exploring  automatic generation of these local ROIs  if the gold-standard ROIs show promising results.

\begin{figure}[t]
\includegraphics[width=0.98\textwidth]{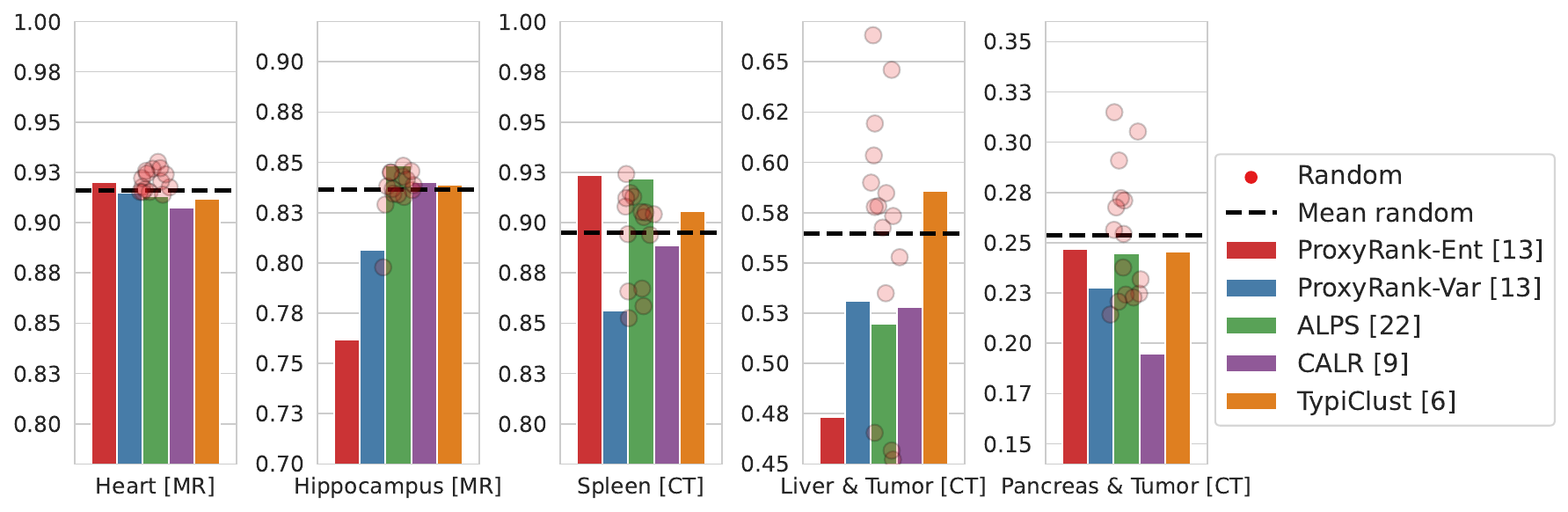}
\centering
\caption{Cold-start AL strategies in a low-budget regime ($m=5$).  TypiClust (orange) is comparable or superior to mean random selection, and consistently outperforms the poor random selection samples. Comprehensive tables are provided in Supp.\ Materials.}
\label{fig2}
\end{figure}

\section{Experimental Results}
\textbf{\underline{Impact of Selection Strategies.}} In Fig. \ref{fig2}, with a fixed budget of 5 samples (except for Heart, where 3 samples are used), we compare the uncertainty-based and diversity-based strategies against the random selection on five different segmentation tasks. Note that the selections made by each of our evaluated AL strategies are deterministic. For random selection, we visualize the individual Dice scores (red dots) of all 15 runs as well as their mean (dashed line). HD95 results (Supp.\ Tab.\ 1) follow the same trends.

Our results explain why random selection remains a strong competitor for 3D segmentation tasks in cold-start scenarios, as no strategy evaluated in our benchmark \textit{consistently} outperforms the random selection \textit{average} performance. 

However, we observe that TypiClust (shown as orange) achieves comparable or superior performance compared to random selection across all tasks in our benchmark, whereas other approaches can significantly under-perform on certain tasks, especially challenging ones like the liver dataset. Hence, \textbf{TypiClust stands out as a more robust cold-start selection strategy}, which can achieve at least a comparable (sometimes better) performance against the mean of random selection. We further note that TypiClust largely mitigates the risk of `unlucky' random selection as it \textit{consistently} performs better than the low-performing random samples (red dots below the dashed line). 

\begin{figure}[t]
\includegraphics[width=\textwidth]{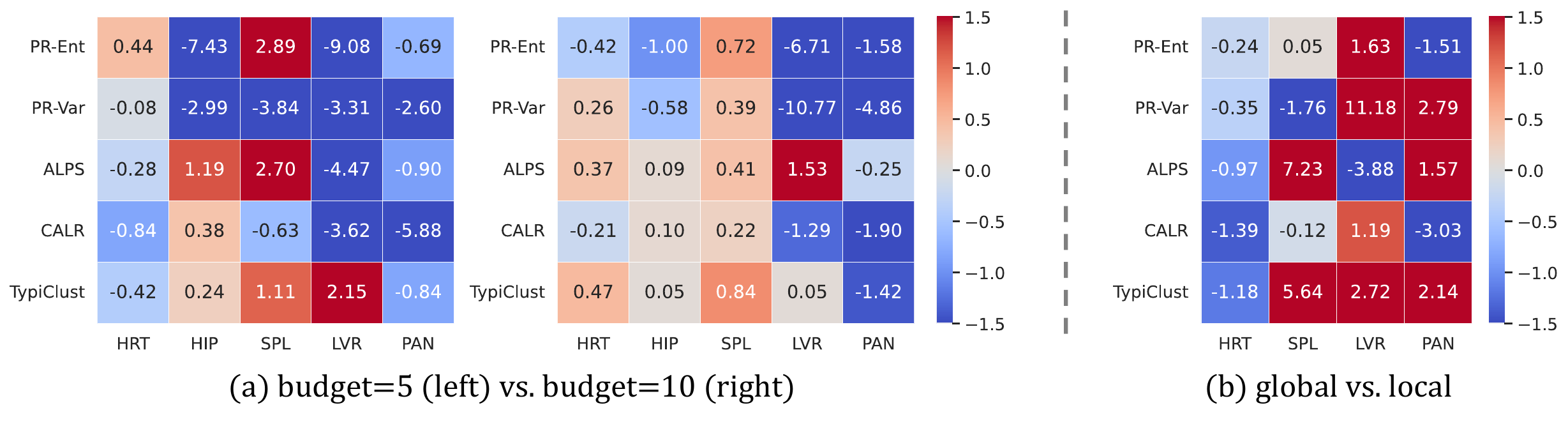}
\centering
\caption{\textbf{(a)} Difference in Dice between each strategy and the mean of the random selection (warm colors: better than random).  Cold-start AL strategies are more effective under the higher budget.  \textbf{(b)} Global vs.\ local ROI performance  (warm colors: global better than local). The local ROI does not yield a consistently better performance. Comprehensive tables are provided in Supp.\ Materials.}
\label{fig3}
\end{figure} 

\noindent
\textbf{\underline{Impact of Different Budgets.}} In Fig.\ \ref{fig3} (a), we compare  AL strategies under the budgets of $m=5$ vs.\ $m=10$ (3 vs.\ 5  for Hearts). We visualize the performance under each budget using a  heatmap, where each element in the matrix is the difference of Dice scores between the evaluated strategy and the mean of random selection under that budget. A positive value (warm color) means that the AL strategy is more effective than random selection. We observe an increasing amount of warm elements in the higher-budget regime, indicating that \textbf{most cold-start AL strategies become more effective when more budget is allowed}. This is especially true for the diversity-based strategies (three bottom rows), suggesting that when a slightly higher budget is available, the diversity of the selected samples is important. HD95 results (Supp.\ Tab.\ 1) are similar.

\noindent
\textbf{\underline{Impact of Different ROIs.}} In Fig.\ \ref{fig3} (b), with a fixed budget of $m=5$ volumes, we compare the AL strategies when uncertainty/diversity is extracted from the entire volume (global) vs.\ a local ROI (local). Each element in this heatmap is the Dice difference of the AL strategy between global and local; warm color means global is better than local. The hippocampus images in MSD are already cropped to the ROI, and thus are excluded from this comparison. We observe different trends across different methods and tasks. Overall, we can observe more warm elements in the heatmap, indicating that \textbf{using only the local uncertainty or diversity for cold-start AL cannot consistently outperform the global counterparts}, even with ideal ROI generated from ground truth. HD95 results (Supp.\ Tab.\ 2) follow the same trends.

\noindent
\textbf{\underline{Limitations.}} For the segmentation tasks that include tumors ($4^{th}$ and $5^{th}$ columns on Fig. \ref{fig3} (a)), we find that almost no AL strategy is very effective, especially the uncertainty-based approaches. The uncertainty-based methods heavily rely on the uncertainty estimated by the network trained on the proxy tasks, which likely makes the uncertainty of tumors difficult to capture. It may be necessary to allocate more budget or design better proxy tasks to make cold-start AL methods effective for such challenging tasks. Lastly, empirical exploration of cold-start AL on iterative AL is beyond the scope of this study and merits its own dedicated study in future.


\section{Conclusion}
In this paper, we presented the COLosSAL benchmark for cold-start AL strategies on 3D medical image segmentation using the public MSD dataset. Comprehensive experiments were performed to answer three important open questions for cold-start AL. While cold-start AL remains an unsolved problem for 3D segmentation, important trends emerge from our results; for example, diversity-based strategies tend to benefit more from a larger budget. Among the compared methods, TypiClust \cite{hacohen2022active} stands out as the most robust option for cold-start AL in medical image segmentation tasks. We believe our findings and the open-source benchmark will facilitate future cold-start AL studies, such as the exploration of different uncertainty estimation/feature extraction methods and evaluation on multi-modality datasets.

\section{Acknowledgements}
This work was supported in part by the National Institutes of Health grants R01HD109739
 and T32EB021937, as well as National Science Foundation grant 2220401. This work was also supported by the Advanced Computing Center for Research and Education (ACCRE) of Vanderbilt University.

\bibliographystyle{splncs04}
\bibliography{references.bib}
\end{document}